\pdfoutput=1

\documentclass[11pt]{article}

\usepackage[]{acl}

\usepackage{times}
\usepackage{latexsym}
\usepackage{float}
\usepackage{tabularx}         
\usepackage{booktabs}         
\usepackage{multirow,multicol}         
\usepackage{graphicx}
\usepackage{cuted}
\usepackage{amsmath}
\usepackage{amssymb}
\usepackage{adjustbox}
\newcommand*{\Perm}[2]{{}^{#1}\!P_{#2}}%

\usepackage[T1]{fontenc}

\usepackage[utf8]{inputenc}

\usepackage{microtype}

\title{Prompt-based Learning for Text Readability Assessment}
\renewcommand*{\thefootnote}{\fnsymbol{footnote}}

\newcommand*{\email}[1]{\texttt{#1}}
\author{
Bruce W. Lee$^{1,2,}$\footnote[3]{}\hspace{1ex}, \ 
Jason Hyung-Jong Lee$^{2}$\ 
\\ 
$^{1}$University of Pennsylvania - PA, USA \\
$^{2}$LXPER AI Research - Seoul, South Korea \\
 \email{brucelws@seas.upenn.edu} \\
 \email{jasonlee@lxper.com} \\}

\begin{document}
\maketitle

\begin{abstract}
We propose the novel adaptation of a pre-trained seq2seq model for readability assessment. We prove that a seq2seq model -- T5 or BART -- can be adapted to discern which text is more difficult from two given texts (pairwise). As an exploratory study to prompt-learn a neural network for text readability in a \textit{text-to-text} manner, we report useful tips for future work in seq2seq training and ranking-based approach to readability assessment. Specifically, we test nine input-output formats/prefixes and show that they can significantly influence the final model performance.

Also, we argue that the combination of text-to-text training and pairwise ranking setup 1) enables leveraging \textit{multiple} parallel text simplification data for teaching readability and 2) trains a neural model for the general concept of readability (therefore, better cross-domain generalization). At last, we report a 99.6\% pairwise classification accuracy on Newsela and a 98.7\% for OneStopEnglish, through a joint training approach. Our code is available at \texttt{github.com/brucewlee/prompt-lea\\rning-readability}.
\end{abstract}

\section{Introduction}
\footnotetext[3]{Woong Sung (Bruce) Lee was on leave from the University of Pennsylvania during the research period.}
\renewcommand*{\thefootnote}{\arabic{footnote}}

Readability assessment evaluates the reading difficulty of a given piece of text \citep{vajjala2021trends}. The early traditional readability assessment methods like Flesch-Kincaid Grade Level \citep{kincaid1975derivation} utilized a linear regression formula fitted to data from large-scale reading experiments on human subjects. More recently, readability assessment has often been viewed as a classification task \citep{feng:2010}. Under this classification-based task formulation, models using various handcrafted features \citep{xia-etal-2016-text, vajjala-meurers-2012-improving}, computer-generated features \citep{Martinc:21, imperial2021bert}, or both \citep{lee-etal-2021-pushing} have been reported. Showing the potential that neural modeling is more suitable than handcrafted features in holistically capturing the inherent linguistic properties that affect readability.

Among the varying approaches to readability assessment, fine-tuning deep transformer models \citep{NIPS2017_3f5ee243}, that are pre-trained with language modeling objectives (e.g. \textit{Masked Language Modelling}, \textit{Next Sentence Prediction}), has proven highly effective in multiple reports \citep{lee2022neural, lee-etal-2021-pushing}. So far, encoder-only transformer architectures like BERT \citep{bert} and RoBERTa \citep{liu2019roberta} have been the go-to approach, and few reports discuss the applicability of other architecture types. Further, there is no report on how readability assessment can be cast in a text-to-text task formulation (\S2). But recent reports \citep{raffel2020exploring} show that a text-to-text is promising for multiple downstream tasks.

The main contribution of this paper is that we fine-tune full encoder-decoder transformer architectures (also referred to as \textit{sequence-to-sequence}) and check if they can learn about text readability. A sequence-to-sequence model has been previously fine-tuned for downstream tasks like document ranking \citep{nogueira2020document}, but few reports discuss whether the architecture can learn about readability.

We fine-tune BART \citep{lewis-etal-2020-bart} and T5 \citep{raffel2020exploring} on the popular OneStopEnglish \citep{Vajjala:18} and Newsela \citep{xu2015problems} data. Then, we measure their performance on two other datasets with readability annotations for US and CEFR curricula, respectively.

We also conduct methodological explorations on how a sequence-to-sequence model can be well-trained to learn text readability. This includes how input and output format should be structured, considering that the fine-tuning for a sequence-to-sequence model is naturally cast in a text-to-text format \citep{nogueira2020document}. We summarize our research questions in two:
\vspace{-2mm}
\begin{itemize}
    \item [1)] Can a sequence-to-sequence model be fine-tuned for text readability -- with a parallel text simplification dataset?
    \vspace{-2mm}
    \item [2)] If so, can the performance generalize across domains? In other words, is the model learning the \textit{dataset} or the concept of \textit{readability}?
\end{itemize}
\vspace{-2mm}

\begin{table*}[ht]
\begin{center}
\footnotesize
\begin{tabular}{lcc}
\toprule
\textbf{Type} & \textbf{Input Format} & \textbf{Target Output}\\ 

\midrule
Question            & "Which Text is more difficult? Text 1: ... Text 2: ..." & "Text 1" or "Text 2"\\
Statement             & "Text 1 is more difficult than Text 2. Text 1: ... Text 2: ..."      & "True" or "False"\\
Follow-up             & "Text 1: ... Text2: ... More difficult:" & "Text 1" or "Text 2"\\
\midrule
Reverse-Question            & "Which Text is easier? Text 1: ... Text 2: ..." & "Text 2" or "Text 1"\\
Reverse-Statement             & "Text 1 is easier than Text 2. Text 1: ... Text 2: ..."      & "False" or "True"\\
Reverse-Follow-up             & "Text 1: ... Text2: ... Easier:" & "Text 2" or "Text 1"\\
\midrule
Alternate-Question            & "Which Text is harder? Text 1: ... Text 2: ..." & "Text 1" or "Text 2"\\
Alternate-Statement             & "Text 1 is harder than Text 2. Text 1: ... Text 2: ..."      & "True" or "False"\\
Alternate-Follow-up             & "Text 1: ... Text2: ... Harder:" & "Text 1" or "Text 2"\\
\bottomrule

\end{tabular}
\caption{Input-Output format candidates we tested. The text-to-text formulation is intuitive  internally (model) and externally (human) because the model input and output are both representations of some semantic concept.}
\end{center}
\vspace{-4mm}
\end{table*}

These research questions and our task approach are intentionally formulated in reference to what previous literature \citep{vajjala2021trends, lee-etal-2021-pushing} have proposed as future directions. As we elaborate further in the following sections, our approach simplifies some inherent problems that we had in the readability assessment. But our study has limitations and requires further explorations (\S5).

\section{Background Knowledge}
\subsection{Sequence-to-Sequence Transformers}
Pre-trained sequence-to-sequence transformers essentially incorporate both encoder and decoder parts from the original Transformer \citep{NIPS2017_3f5ee243} architecture. The most notable examples are T5 and BART, pre-trained using document denoising strategies (i.e. during pre-training, input is an intentionally corrupted document, and output is a recovered document or corrupted portions). 

Though BART allows some flexibility in altering the model architecture for downstream tasks \citep{lewis-etal-2020-bart}, T5 is built with the intention of unifying all NLP tasks into a text-to-text-format \citep{raffel2020exploring}. Here, text-to-text means that both input and output are always text strings, unlike encoder-only, BERT-style models that only output either a class label ([CLS] token) or an input span.

\subsection{Existing Downstream Tasks}
Pre-trained sequence-to-sequence models can be fine-tuned to most NLP downstream tasks, including neural machine translation \citep{wang2022understanding} and abstractive summarization \citep{saito2020abstractive}. Then, a training instance is formatted in a way that is much like \textit{telling} the model what to do by adding task-specific prefix \citep{raffel2020exploring}. 

If a model were to be fine-tuned for translation, the input format could be “\textit{translate English to German: That is good.}” and the target output is “\textit{Das ist gut.}” When applied to document ranking, \citet{nogueira2020document} proposed a slightly different format. The input format was written "\textit{Query: ... Document: ... Relevant:}" so that the target output tokens -- "\textit{true}" or "\textit{false} -- can naturally come after the input format. Our work is the most influenced by this formatting approach.

\section{Experimental Setup}
\subsection{Methods}
All our experiments are based on T5 and BART, both obtained from the respective online repositories through Huggingface \citep{wolf2019huggingface}. Since the text-to-text formulation has never been used to fine-tune text readability\footnote{\citet{lee-etal-2021-pushing} trains BART for readability but uses class labels (BERT-style) as target output, instead of text tokens}, here we limit to the simple task of comparing the readability of two given texts. Following our input format (Table 1), we fed two text snippets of varying difficulties to the model every instance. Then, the model was trained to give the corresponding target output. We tested nine input-output formats, as shown in Table 1.

\subsection{Datasets and Preparation}
\subsubsection{Data Type and Permutation Methods}
The datasets that we use in this study are of two types. \textit{Parallel} type contains multiple reading-level versions of a text (mostly through human expert paraphrasing). We call a grouping of text in multiple reading levels a \textit{slug}. On the other hand, there is \textit{distinct} type of datasets. \textit{Distinct} type is a more common format where each text is given a readability level with no multiple versions of the same content. Our naming and permutation strategies are inspired by existing work on pairwise ranking for readability \citep{lee2022neural}.

A parallel dataset $D_{parallel}$ can be expressed as a row-wise collection of $i$ slugs $D_{parallel} = [S_1,...,S_i]$, where a slug is a column-wise collection of $j$ pairs of a text and a reading level $S_i = [(x_{i,1}, y_{i,1}),...,(x_{i,j}, y_{i,j})]$. For parallel dataset, we perform permutation $\Perm{j}{2}$ on the slug level, creating $S_i^{'} = [((x_{i,1}, x_{i,2}),(y_{i,1}, y_{i,2})),...,((x_{i,j-1}, x_{i,j}),\\(y_{i,j-1}, y_{i,j}))]$. A pair like $((x_{i,1}, x_{i,2}),(y_{i,1}, y_{i,2}))$ is considered an instance for train/dev/test. Then, \\$D_{parallel}^{'} = [S_1^{'},...,S_i^{'}]$, where all $S^{'}$ are flattened to make $D_{parallel}^{'}$ an iterable collection of tuples. This setup is intended to be robust to future implementations of paraphrase-based text simplification datasets where the standards for readability annotation/level are only consistent within a slug.

A distinct dataset $D_{distinct}$ can be expressed as a collection of $i$ pairs of a text and a reading level $D_{distinct} = [(x_{1}, y_{1}),...,(x_{i}, y_{i})]$. Then, we can perform permutation $\Perm{i}{2}$ to create $D_{distinct}^{'} = [((x_{1}, x_{2}),(y_{1}, y_{2})),...,((x_{i-1}, x_{i}),(y_{i-1}, y_{i}))]$. This setup is under the postulation that readability annotation is consistent throughout the dataset. Pairwise instances of two same levels are removed.

\subsubsection{Datasets}
Two \textit{parallel} and two \textit{distinct} datasets are used. Also, we agree with \citet{vajjala2021trends}'s concern that "\textit{(in some datasets) articles tagged with different reading levels don’t share the same topical content (...) question what the readability assessment models learn - is it a notion of text complexity, or topical differences among texts?}". Hence, we only use parallel data -- NEWS and OSEN -- for training.

Our dataset processing strategy (\S3.2.1) and pairwise comparison approach force the model to learn label-agnostic, the global concept of relative difficulties of texts. That is, the model learns that a text annotated level 5 should be harder than a level 3 (or a level 2 or a level 1) within a slug or a dataset. Such a setup is inherently robust against cross-domain usage (Table 3). Further, it enables combining multiple datasets of various slug sizes or readability annotation standards for joint training (\S4).

\textbf{Newsela}(NEWS) is a \textit{parallel} text simplification dataset introduced by \citet{xu2015problems}. It consists of 2,154 slugs, each item re-written 4 or 5 times for children at different grade levels. Hence, a total of 10,786 texts are contained, and 43,316 pairwise instances are created after data permutation (\S3.2.1). Random shuffling split these instances into 6:2:2 for train/test splits. We provide reproducible scripts for all datasets through code.

\begin{table}[t]
    \centering
    \footnotesize
    \begin{tabular}{l|c@{\hspace{0.8ex}}c@{\hspace{0.8ex}}c@{\hspace{0.8ex}}c}
    \toprule
    \multirow{2.5}{*}{Format Type} & \multicolumn{2}{c|}{\textbf{OSEN}} &\multicolumn{2}{c}{\textbf{NEWS}} \\
    \cmidrule{2-5}
    & \textbf{T5}&\textbf{BART}& \textbf{T5}&\textbf{BART} \\
    \midrule
    Question     &\textbf{0.815(30)}  &0.965(18)   &\textbf{0.981(3)}  &-\\
    Statement    &0.639(29)  &0.978(30)   &-      &-\\
    Follow-up    &0.784(25)  &0.960(27)   &-      &-\\
    Reverse-Q    &0.793(30)  &\textbf{0.991(30)}   &-      &\textbf{0.993(3)}\\
    Reverse-S    &0.524(28)  &0.978(26)   &-      &-\\
    Reverse-F    &0.828(30)  &\textbf{0.991(30)}   &-      &\textbf{0.993(5)}\\
    Alternate-Q  &0.811(30)  &0.960(25)   &-      &-\\
    Alternate-S  &0.617(29)  &0.978(30)   &-      &-\\
    Alternate-F  &0.789(28)  &0.938(29)   &-      &-\\
    \bottomrule
    \end{tabular}
\caption{Validation set accuracy reports on NEWS and OSEN. The best epoch is reported in brackets. NEWS is only reported for the best format type due to data size.}
\vspace{-4mm}
\end{table}

\textbf{OneStopEnglish}(OSEN) is a \textit{parallel} dataset intended for both text simplification and readability assessment research \citep{Vajjala:18}. It consists of 189 slugs, each item in 3 paraphrases at different reading levels. A total of 567 texts are contained, and 1,134 pairwise instances are created. We use a 6:2:2 split ratio through random shuffling.

\textbf{Common Core Standards}(CCSB) is a \textit{distinct} collection of exemplary official texts with readability annotations in U.S. Common Core Standards. We scraped data from the source ourselves. We used 69 story-type texts in 6 reading levels. After permutation, 3,846 pairwise instances are created. 

\textbf{Cambridge English Readability}(CAMB) is a \textit{distinct} dataset of reading passages from main suite Cambridge English Exams \citep{xia-etal-2016-text}. All 331 texts are labeled A2, B1, B2, C1, or C2 reading levels, following the CEFR standards. After permutation, 87,574 pairwise instances are created. 

\subsection{Training}
The batch size is fixed at 8, both for training and inference. The learning rate is fixed at 1e-5 for T5 and BART. We fine-tune OSEN for 30 epochs and NEWS for 3 epochs. We report the best epoch performance based on the validation set in Table 2. For joint training (Table 3), we take an OSEN-trained model and then fine-tune further using NEWS for 3 more epochs.

\section{Results}
1. \textbf{A pretrained sequence-to-sequence model could be fine-tuned for text readability, in a text-to-text style.} Table 2 shows that the concept of readability could be fine-tuned in a text-to-text task formulation, some setups with decent accuracies of $>0.9$. For a smaller dataset (OSEN), BART significantly outperformed T5, but their performance deviation was little on a larger dataset (NEWS). We believe this is caused due to difference in pre-training methodologies that caused T5 to require more training steps to learn about our downstream task. Also, BART always generalized better than T5 across unseen datasets in Table 3.

2. \textbf{Input-output format significantly affected the final performance, especially when fine-tuning T5 with lesser training steps (OSEN).} Among the nine input-output formats we tested, T5 and BART performed best under \texttt{Question} and \texttt{Reverse-Q/F} types, respectively. Performance deviations caused by input-output format changes were larger than we expected. Further, no certain format generally ensured good results across models. This raises the need for additional "format-tuning" processes when exploring new models. However, it must be noted that several observations point to T5 being under-trained for the general concept of readability at the data size of OSEN (see Table 2 and Table 3). The input-output format's influence is lesser for setups where models learned better about readability.

\begin{table}[t]
    \centering
    \begin{adjustbox}{max width=0.48\textwidth}
    \footnotesize
    \begin{tabular}{l@{\hspace{0.8ex}}|c@{\hspace{0.8ex}}c@{\hspace{0.8ex}}c@{\hspace{0.8ex}}c}
    \toprule
    \multirow{2.4}{*}{\textbf{Model / Fine-Tune Data}} & \multicolumn{4}{c}{\textbf{Test Data}}\\
    \cmidrule{2-5}
    &\textbf{OSEN} & \textbf{NEWS} & \textbf{CCSB} & \textbf{CAMB} \\
    \midrule
    Flesch-Kincaid / None&0.978            &0.986         &\textbf{0.798}&0.808\\
    T5 / OSEN            &\underline{0.784}&\textit{0.518}&\textit{0.509}&\textit{0.492}\\
    BART / OSEN          &\underline{0.978}&\textit{0.871}&\textit{0.639}&\textit{0.629}\\
    T5 / NEWS            &\textit{0.907}&\underline{0.967}&\textit{0.747}&\textit{0.764}\\
    BART / NEWS          &\textbf{\textit{0.987}}&\underline{0.993}&\textit{0.793}&\textit{\textbf{0.883}}\\
    T5 / OSEN + NEWS     &\underline{0.992}&\underline{0.987} &\textit{0.771}&\textit{0.778}\\
    BART / OSEN + NEWS   &\underline{0.983}&\textbf{\underline{0.996}}&\textit{0.790}&\textit{0.865}\\
    \bottomrule
    \end{tabular}
    \end{adjustbox}
\caption{\underline{In-domain} and \textit{cross-domain} accuracies across datasets. For OSEN and NEWS, test sets (\S3.2.2) are used. The best result per dataset is in \textbf{bold}. \texttt{T5} is trained with \texttt{Question} format, whereas \texttt{BART} is trained with \texttt{Reverse-F} format. Flesch-Kincaid refers to the popular Flesch-Kincaid Grade Level formula published in \citet{kincaid1975derivation}. We use the implementation in \texttt{github.com/textstat/textstat}.}
\vspace{-4mm}
\end{table}

3. \textbf{Joint training has the potential to help both in-domain and cross-domain performances.} Joint training of multiple datasets for a single model is an under-explored concept in readability assessment. This is because human experts annotate existing datasets with varying standards. Dataset construction can also differ (e.g. different number of classes or too difficult to map classes). Hence, it was unknown if combining datasets of varying labeling standards improves performance.

This work solves the problem by re-casting the task into a simple, universal question of comparing two texts' difficulties (\S3.2.2). Table 3 shows that in- and cross-domain performances can improve through joint training. For example, in-domain accuracies for OSEN increased to 0.208 when the model was further fine-tuned with a larger extra data, NEWS. However, a NEWS-only model generally performed better than the OSEN+NEWS model in Table 3. We expect that OSEN, which is almost 40 times smaller, only confused the model.

4. \textbf{Exposing the model to more texts with a wider range of readability helped fine-grained readability comparison.} Importantly, we showed that readability assessment models fine-tuned with \textit{parallel} datasets could be generalized across \textit{distinct} datasets (e.g. OSEN $\rightarrow$ CCSB). But model performances varied depending on label distance. Models performed better when the two compared texts' readability labels were larger apart (i.e. the model is more likely to guess level 1 vs level 4 correctly than level 1 vs level 2). This problem worsened when the model was trained using OSEN. Using NEWS as training data or extra data helped. We want to point out that a slug size in NEWS is 4 or 5, exposing the model to more permutations.

5. \textbf{Text-to-text style fine-tuning required more training steps than expected.} The majority of our OSEN fine-tuning experiments showed that the model's validation set performance continues to increase up to epoch 30. This is contrastive to how usual classification approaches, using encoder-only models, only fine-tune up to epoch 3$\sim$5 even on smaller datasets like OSEN or CAMB \citep{lee-etal-2021-pushing}. Intuitively speaking, there is potential that better performance can be achieved if fine-tuned further. We will explore this concept in the future.

6. \textbf{Though often overlooked, traditional readability formulas provide challenging baselines.} The traditional readability formulas are criticized for their low performances in multi-class ranking or regression-based readability task formulation \citep{lee2023traditional}. However, they provide surprisingly strong baselines for pairwise difficulty comparisons, as seen in Table 3.

\section{Conclusion}
So far, we have reported our exploratory work on casting readability assessment tasks in a text-to-text formulation for BART and T5. We summarized our observations into five categories in \S4, which can serve as base guidelines for future work. Our experimental setup and data permutation methods allow the joint training of more than one dataset, regardless of whether the dataset construction is \textit{parallel} or \textit{distinct} (\S3). Using NEWS as extra training data further to fine-tune an OSEN-trained model greatly improved model performance. However, we did not train the other way around (NEWS $\rightarrow$ OSEN), which should be proved in the future.

\section{Limitations}
Our limitations are in input text length and output labels. Though our novel task formulation allows the application of essential concepts like \textit{joint training} or \textit{cross-domain evaluation} in the field of readability assessment, it is based on a pairwise classification method. Since the pairwise approach only allows the readability ranking of two texts (e.g. which is easier?), it lacks practicality compared to regression or multi-label classification-based models. Though we achieve an almost perfect accuracy of 99.6\% in Newsela data, knowing which is easier out of texts has little use as a real-world system. Hence, further research must be conducted to generate more useful output labels and process longer sequences. Like \citet{nogueira2020document}, we are looking into using a sliding window to generate output labels for longer input sequences. We are also researching neural models pre-trained specifically for readability assessment using the prompt-based learning method introduced in the paper. Such a model can be leveraged for multi-class classification.


\bibliography{anthology,custom}
\bibliographystyle{acl_natbib}
\newpage
\appendix

\section{Obtaining Dataset}
We obtained Newsela by requesting academic access at \textit{newsela.com/data}. 

\noindent OneStopEnglish dataset is freely available at \textit{github.com/nishkalavallabhi/OneStopEnglishCorpus}.

\noindent Cambridge English Readability dataset is freely available at \textit{ilexir.co.uk/datasets/index.html}.

\noindent We crawled Common Core Appendix B from \textit{www.corestandards.org/assets/Appendix\_B.pdf}.

\section{Training Details}
\subsection{Models, GPU, Train Time}
All pre-trained models are retrieved from Huggingface. Fine-tuning code is written in PyTorch. We used a single NVIDIA RTX 2080 GPU for all our training.

\noindent\textbf{T5}
\begin{itemize}
\item huggingface.co/t5-base

\item 1,020 seconds for 30 epochs on permutated OneStopEnglish, in \texttt{Question} format

\item 6,994 seconds for 3 epochs on permutated Newsela, in \texttt{Question} format
\end{itemize}

\noindent\textbf{BART}
\begin{itemize}
\item huggingface.co/facebook/bart-base

\item 532 seconds for 30 epochs on permutated OneStopEnglish, in \texttt{Reverse-F} format

\item 4,084 seconds for 3 epochs on permutated Newsela, in \texttt{Reverse-F} format
\end{itemize}

\subsection{More on Input Sequence Length}
The current experimental setup is inherently weak against long texts. This is because we input a set of two texts with format prefixes (Table 1). Hence, for a model with 512 max token length our actual token length limitation per text is $\leq256$, with the exact number depending on chosen format type. Though both BART and T5 support longer input sequences, in comparison to other popular models like BERT or RoBERTa, we must empirically confirm if good performance can be achieved when BART and T5 are used with longer max sequence length. In this paper, the max sequence length was set to 512 (which means that $\leq256$ token limit existed per text), and most texts from OSEN had to be truncated before training.

\end{document}